\documentclass[11pt,a4paper,acceptedWithA]{article}
\usepackage{times,latexsym}
\usepackage{url}
\usepackage[T1]{fontenc}

\usepackage[]{tacl2021v1}

\usepackage{xspace,mfirstuc,tabulary}

\usepackage[utf8]{inputenc}

\usepackage{microtype}

\usepackage{amsfonts}
\usepackage{inconsolata}
\usepackage{booktabs}
\usepackage{enumitem}
\usepackage{algorithm,algorithmicx}
\usepackage[noend]{algpseudocode}
\usepackage{multirow}
\usepackage{color}
\usepackage{bbding}
\usepackage{graphicx}
\usepackage{amsmath}
\usepackage{xcolor}
\usepackage{colortbl}
\usepackage{arydshln}
\usepackage{utfsym}
\usepackage{fontawesome }

\newif\iftaclinstructions
\taclinstructionsfalse 
\iftaclinstructions
\renewcommand{\confidential}{}
\renewcommand{\anonsubtext}{(No author info supplied here, for consistency with
TACL-submission anonymization requirements)}
\newcommand{\instr}
\fi

\iftaclpubformat 

\else

\fi

\title{Is Sarcasm Detection A Step-by-Step Reasoning Process in Large Language Models?}

\author{
  \bf Ben Yao$^a$, Yazhou Zhang$^{b,c}$, Qiuchi Li$^a$, Jing Qin$^b$
  \\
  \ \\
  $^a$University of Copenhagen, $^b$The Hong Kong Polytechnic University, $^c$Tianjin University
  \\
}

\date{}

\begin{document}

\maketitle

\begin{abstract}
Elaborating a series of intermediate reasoning steps significantly improves the ability of large language models (LLMs) to solve complex problems, as such steps would evoke LLMs to think sequentially. However, human sarcasm understanding is often considered an intuitive and holistic cognitive process, in which various linguistic, contextual, and emotional cues are integrated to form a comprehensive understanding, in a way that does not necessarily follow a step-by-step fashion. To verify the validity of this argument, we introduce a new prompting framework (called SarcasmCue) containing four sub-methods, $viz.$ chain of contradiction (CoC), graph of cues (GoC), bagging of cues (BoC) and tensor of cues (ToC), which elicits LLMs to detect human sarcasm by considering sequential and non-sequential prompting methods. 
Through a comprehensive empirical comparison on four benchmarks, we highlight three key findings:
(1) CoC and GoC show superior performance with more advanced models like GPT-4 and Claude 3.5, with an improvement of 3.5\% $\uparrow$.
(2) ToC significantly outperforms other methods when smaller LLMs are evaluated, boosting the F1 score by 29.7\% $\uparrow$ over the best baseline.
(3) Our proposed framework consistently pushes the state-of-the-art (i.e., ToT) by 4.2\%, 2.0\%, 29.7\%, and 58.2\% in F1 scores across four datasets. This demonstrates the effectiveness and stability of the proposed framework.\footnote{Our codes are available at https://github.com/qiuchili/llm\_sarcasm\_detection}.
\end{abstract}

%

\section{Introduction}
 \begin{figure}[t]
    \centering
    \includegraphics[width=3in]{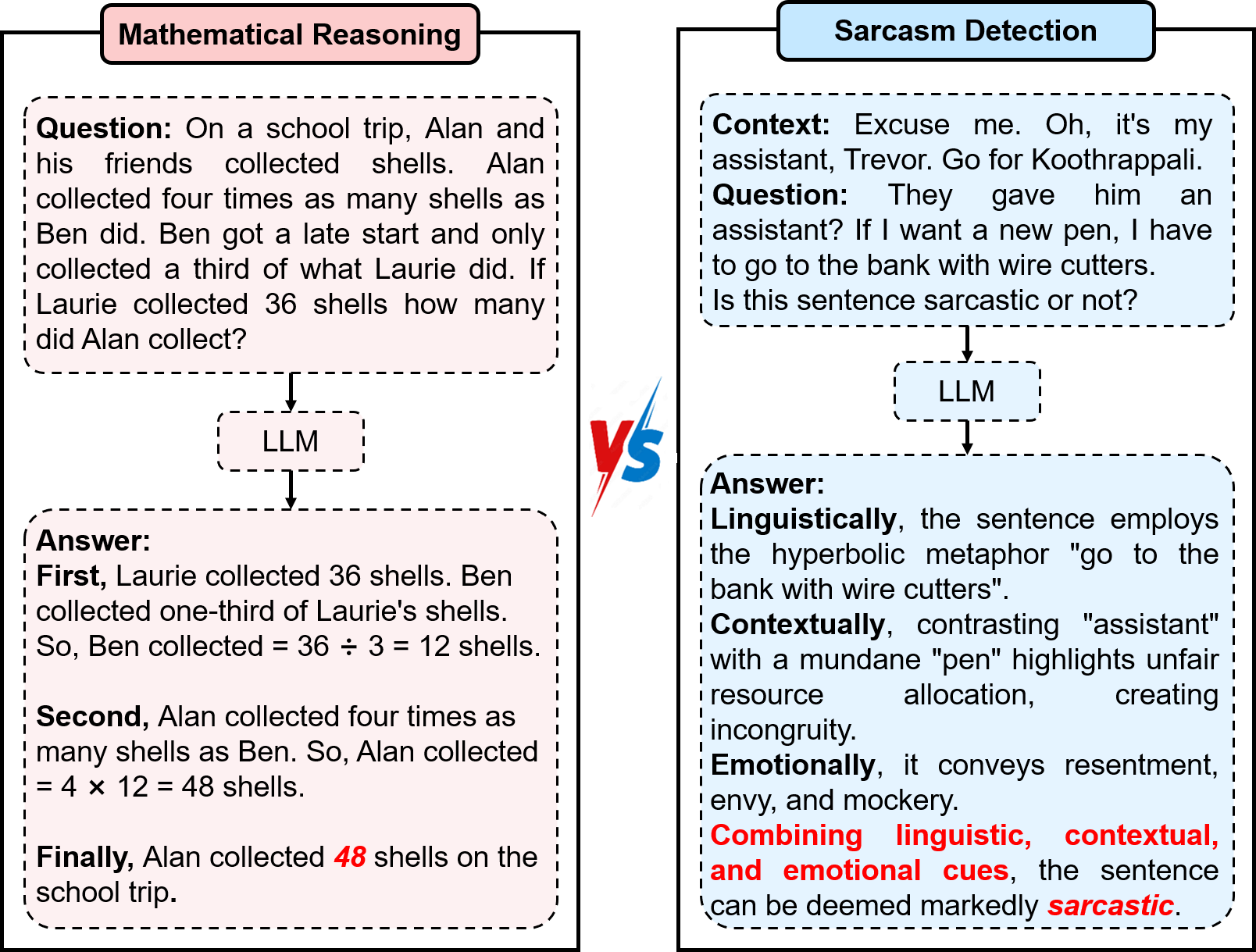}
  \caption{The comparison of the processes of mathematical reasoning and sarcasm detection.}
   \label{fig:example}
\end{figure}

Recent large language models have demonstrated impressive performance across downstream natural language processing (NLP) tasks, in which ``System 1''~-~the fast, unconscious, and intuitive tasks, e.g., sentiment classification, topic analysis, etc., have been argued to be successfully performed~\cite{cui2024survey}. Instead, increasing efforts have been devoted to the other class of tasks~-~``System 2'', which requires slow, deliberative and multi-steps thinking, such as logical, mathematical, and commonsense reasoning tasks~\cite{wei2022chain}. To improve the  ability of LLMs to solve such complex problems, a popular paradigm is to decompose complex problems into a series of intermediate solution steps, and elicit LLMs to think step-by-step, such as chain of thought (CoT)~\cite{wei2022chain}, tree of thought (ToT)~\cite{yao2024tree}, graph of thought (GoT)~\cite{besta2024graph}, etc.

However, due to its inherent ambivalence and figurative nature, sarcasm detection is often considered a holistic and non-rational cognitive process that does not conform to step-by-step logical reasoning for two main reasons: (1) sarcasm expression does not strictly conform to formal logical structures, such as the law of hypothetical syllogism (i.e., \textit{if} $\mathcal{A} \Rightarrow \mathcal{B} $ \textit{and} $\mathcal{B} \Rightarrow \mathcal{C} $, \textit{then} $\mathcal{A} \Rightarrow \mathcal{C} $). For example, ``\textit{Poor Alice has fallen for that stupid Bob; and that stupid Bob is head over heels for Claire; but don't assume for a second that Alice would like Claire}''; (2) sarcasm judgment is often considered a fluid combination of various cues.  Each cue holds equal importance and there is no rigid sequence of steps among them, as shown in Fig.~\ref{fig:example}. Hence, the main research question can be summarized as:

\textbf{RQ:}  \textit{Is human sarcasm detection a step-by-step reasoning process?}

To answer this question, we propose a theoretical framework, called SarcasmCue, based on the sequential and non-sequential prompting paradigm. It consists of four prompting methods, i.e., \textit{chain of contradiction (CoC)}, \textit{graph of cues (GoC)}, \textit{bagging of cues (BoC)} and \textit{tensor of cues (ToC)}. Each method has its own focus and advantages. In this work, \textit{cue} is similar to \textit{thought}, being a coherent language sequence related to linguistics, context, or emotion that serves as an intermediate indicator for identifying sarcasm, such as rhetorical devices or emotional words.
More specifically, 
\begin{itemize}[itemsep=0pt, topsep=0pt, parsep=4pt, leftmargin=*]
	\item{\textbf{CoC.} It harnesses the quintessential property of sarcasm (namely the contradiction between surface sentiment and true intention). It aims to: (1) identify the surface sentiment by extracting keywords, etc.; (2) deduce the true intention by scrutinizing rhetorical devices, etc.; and (3) determine the inconsistency between them. It is a typical linear structure.}
	\item{\textbf{GoC.} Generalizing over CoC, GoC frames the problem of sarcasm detection as a search over a graph and treats various cues as nodes, with the relations across cues represented as edges. Unlike CoC and ToT, it goes beyond following a fixed hierarchy or linear reasoning path. In summary, both CoC and GoC follow the step-by-step reasoning process. }
	\item{\textbf{BoC.} BoC is a bagging approach that constructs a pool of diverse cues and randomly sampling multiple cue subsets. LLMs are employed to generate multiple predictions based on these subsets, and such predictions are aggregated to produce the final result. It is a set-based structure.}
	\item{\textbf{ToC.} ToC treats each type of cues (namely linguistic, contextual, and emotional cues) as an independent, orthogonal view for sarcasm understanding and constructs a multi-view representation through the tensor product. It allows language models to leverage higher-order interactions among the cues. ToC can be visualized as a 3D volumetric structure. Hence, BoC and ToC are proposed based on the assumption that sarcasm detection is not a step-by-step reasoning process.}
\item{\textbf{Their correlation.} These four methods represent an evolution from linear to nonlinear, and from a single perspective to multiple perspectives, together forming a comprehensive theoretical framework (SarcasmCue). Their design aims to adapt to various sarcasm detection scenarios.}
\end{itemize}



We present empirical evaluations of the proposed prompting approaches across four benchmarks over 4 SOTA LLMs (i.e., GPT-4o, Claude 3.5 Sonnet, Llama 3-8B, Qwen 2-7B), and compare their results against 3 SOTA prompting approaches (i.e., standard IO prompting, CoT and ToT).
we highlight three key observations:
(1) When the base model is more advanced (such as GPT-4 and Claude 3.5 Sonnet), CoC and GoC show superior performance against the state-of-the-art (SoTA) baseline with an improvement of 3.5\% $\uparrow$.
(2) ToC achieves the best performance when smaller LLMs are evaluated. For example, in Llama 3-8B, ToC's average F1 score of 65.24 represents a 29.7\% improvement over the best baseline method, ToT.
In Qwen 2-7B, ToC shows a 58.2\% improvement over the best baseline method, IO. 
(3) Our proposed framework consistently pushes SoTA by 4.2\%, 2.0\%, 29.7\% and 58.2\% in F1 scores across four datasets. This demonstrates the effectiveness of the proposed framework.
The main contributions are concluded as follows:
\begin{itemize}
\item Our work is the first to investigate the stepwise reasoning nature of sarcasm detection by using both sequential and non-sequential prompting methods.

\item We propose a new prompting framework that consists of four sub-methods, $viz.$ CoC, GoC, BoC and ToC.

\item Comprehensive experiments over four datasets demonstrate the superiority of the proposed prompting framework.
\end{itemize}


\section{Related Work}

\subsection{Chain-of-Thought Prompting}
Inspired by the step-by-step thinking ability of humans, CoT prompting was proposed to ``prompt'' language models to produce intermediate reasoning steps. Wei et al.~\shortcite{wei2022chain} made a formal definition of CoT prompting in LLMs and proved its effectiveness by presenting empirical evaluations on arithmetic reasoning benchmarks.  However, its performance hinged on the quality of manually crafted prompts. To fill this gap, Auto-CoT was proposed to automatically construct demonstrations with questions and reasoning chains~\cite{zhang2022automatic}. 
Furthermore, Yao et al.~\shortcite{yao2024tree} introduced a non-chain prompting framework, namely ToT, which made LLMs consider multiple different reasoning paths to decide the next course of action. Beyond CoT and ToT approaches, Besta et al.~\shortcite{besta2024graph} modeled the information generated by an LLM as an arbitrary graph (i.e., GoT), where units of information were considered as vertices and the dependencies between these vertices were edges. 

However, all of them adopt the sequential decoding paradigm of  ``let LLMs think step by step''. Contrarily, it is argued that sarcasm judgment does not conform to step-by-step logical reasoning, and there is an urgent need to develop non-sequential prompting approaches.
 \begin{figure*}[t]
    \centering
    \includegraphics[width=6.3in]{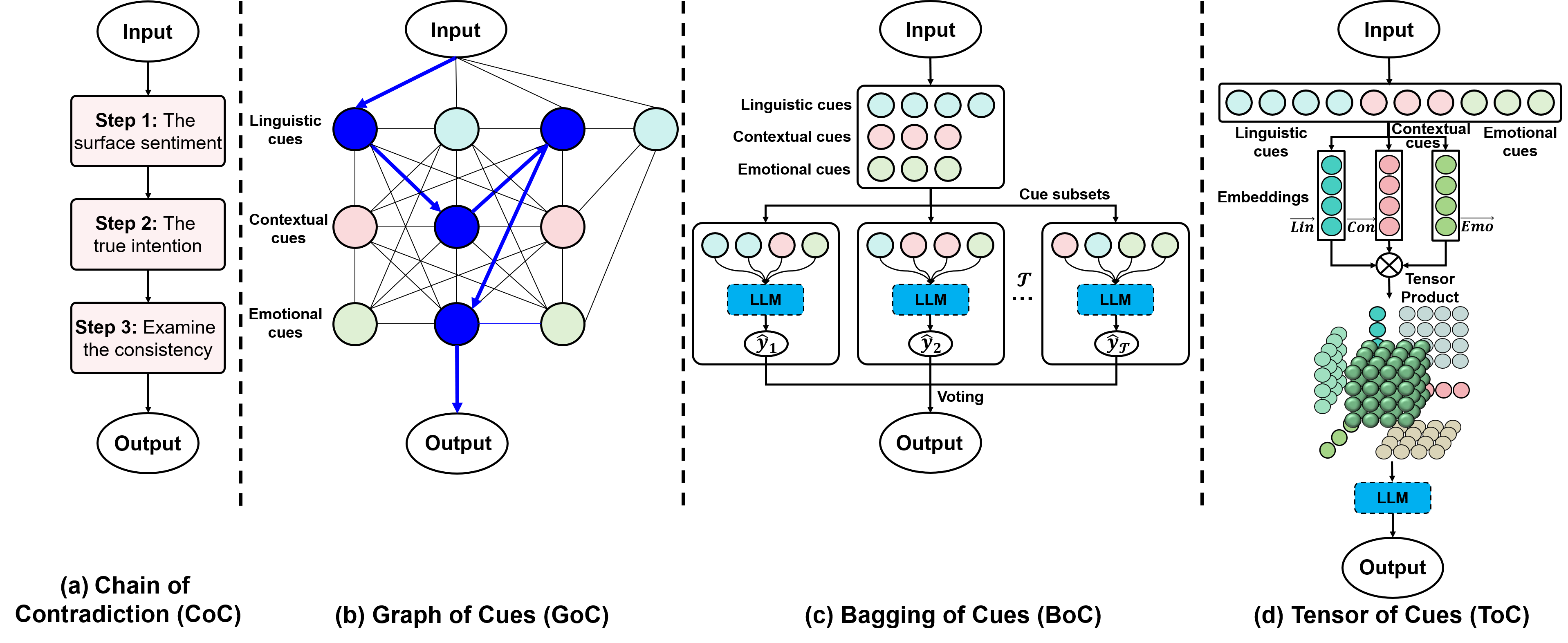}
  \caption{An illustration of our SarcasmCue framework that consists of four prompting sub-methods.}
   \label{fig:model}
\end{figure*}

\subsection{Sarcasm Detection}
Sarcasm detection has evolved from early statistical learning based approaches to traditional neural methods, and further advanced to modern neural methods epitomized by Transformer models. In early stage, statistical learning based approaches mainly employ statistical learning techniques, e.g., SVM, NB, etc., to extract patterns and relationships within the data~\cite{zhang2023stance}.
As deep learning based architectures have shown the superiority, numerous base neural networks, e.g., such as CNN~\cite{jain2020sarcasm}, LSTM~\cite{ghosh2018sarcasm}, GCN~\cite{liang2022multi}, etc., have been predominantly utilized during the middle stage of sarcasm detection research. Now, sarcasm detection research has stepped into the era of pre-trained language models (PLMs). An increasing number of researchers are designing sophisticated PLM architectures to serve as encoders for obtaining effective text representations~\cite{liu2023prompt}. 

Different from them, we propose four prompting methods to make the first attempt to explore the potential of prompting LLMs in sarcasm detection.

\section{The Proposed Framework: SarcasmCue}
The proposed SarcasmCue framework is illustrated in Fig.~\ref{fig:model}. We qualitatively compare SarcasmCue with other prompting approaches in Tab.~\ref{tab:comparison}. SarcasmCue is the only one to fully support chain-based, tree-based, graph-based, set-based and multidimensional array-based reasoning. It is also the only one that simultaneously supports both sequential and non-sequential prompting methods.
\begin{table}[t]
\centering
\footnotesize
\caption{Comparison of prompting methods.} 
\label{tab:comparison}
\scalebox{0.85}{
\begin{tabular}{lccccc}
\toprule
\multirow{2}{*}{\textbf{Scheme}}  & \multicolumn{3}{c}{\textbf{Seq?}} &  \multicolumn{2}{c}{\textbf{Non-Seq?}} \\
\cline{2-6}

&      \textbf{Chain?}      & \textbf{Tree?} & \textbf{Graph?} & \textbf{Set?} & \textbf{Tensor?} \\
\toprule
IO   & \faTimes & \faTimes &  \faTimes  & \faTimes & \faTimes \\

CoT & \faCheckSquare & \faTimes &  \faTimes  & \faTimes & \faTimes \\

ToT & \faCheckSquare & \faCheckSquare &  \faTimes  & \faTimes & \faTimes \\

GoT & \faCheckSquare & \faCheckSquare &  \faCheckSquare  & \faTimes & \faTimes \\\toprule

\textbf{SarcasmCue} & \faCheckSquare & \faCheckSquare &  \faCheckSquare  & \faCheckSquare & \faCheckSquare \\
\toprule
\end{tabular}
}
\end{table}

\subsection{Task Definition}
Given the data set $\mathcal{D}=\left \{ \left ( \mathcal{X}, \mathcal{Y} \right )  \right \}$, where $\mathcal{X} = \{x_1, x_2, \ldots, x_{n}\}$ denotes the input text sequence and $\mathcal{Y} = \{y_1, y_2, \ldots, y_{n}\}$ denotes the output label sequence. We use $\mathcal{L}_{\theta} $ to represent a large language model with parameter $\theta$. Our task is to leverage a collection of cues $\mathcal{C}=\{c_1, c_2,...,c_k\}$ to brige the input $\mathcal{X}$ and the output $\mathcal{Y}$, where each cue $c_i$ is a coherent language sequence that serves as an intermediate indicator toward identifying sarcasm.

\subsection{Chain of Contradiction}
We capture the inherent paradoxical nature of sarcasm, which is the incongruity between the surface sentiment and the true intention, and propose \textit{chain of contradiction}, a CoT-style paradigm that allows LLMs to decompose the problem of sarcasm detection into intermediate steps and solve each before making decision (Fig.~\ref{fig:model} (a)). Each cue $c_k\sim \mathcal{L}_{\theta}^{CoC}\left ( c_k| \mathcal{X},c_{1},c_2,...,c_{k-1}\right ) $ is sampled sequentially, then the output $\mathcal{Y}\sim \mathcal{L}_{\theta}^{CoC}\left ( \mathcal{Y}| \mathcal{X},c_{1},...,c_k\right ) $. A specific instantiation of CoC involves three steps:

\textbf{Step 1.} We first ask LLM to detect the surface sentiment via the following prompt $p_1$:

\noindent \colorbox{gray!13}{\parbox{7.5cm}{Given the input sentence [$\mathcal{X}$], what is the SURFACE sentiment, as indicated by clues such as keywords, sentimental phrases, emojis?}}

\noindent $c_1$ is the output sequence, which can be formulated as $c_1\sim \mathcal{L}_{\theta}^{CoC}\left ( c_1| \mathcal{X},p_1\right ) $.

\textbf{Step 2.} We thus ask LLM to carefully discover the true intention via the following prompt $p_2$:

\noindent \colorbox{gray!13}{\parbox{7.5cm}{Deduce what the sentence really means, namely the TRUE intention, by carefully checking any rhetorical devices, language style, unusual punctuations, common senses.}}

\noindent $c_2$ is the output sequence, which can be formulated as $c_2\sim \mathcal{L}_{\theta}^{CoC}\left ( c_2| \mathcal{X}, c_1, p_2\right ) $.

\textbf{Step 3.} Let LLM examine the consistency between surface sentiment and true intention and make the final prediction:

\noindent \colorbox{gray!13}{\parbox{7.5cm}{Based on Step 1 and Step 2, evaluate whether the surface sentiment aligns with the true intention. If they do not match, the sentence is probably `Sarcastic'. Otherwise, the sentence is `Not Sarcastic'. Return the label only.}}


CoC raises a presumption that the cues are linearly correlated, and detects human sarcasm through step-by-step reasoning. Further details see Algorithm~1 in App. A.

\subsection{Graph of Cues}
The linear structure of CoC restricts it to a single path of reasoning. To fill this gap, we introduce \textit{graph of cues}, a graph based paradigm that allows LLMs to flexibly choose and weigh multiple cues, unconstrained by the need for unique predecessor nodes (Fig.~\ref{fig:model} (b)). GoC frames the problem of sarcasm detection as a search over a graph, and is formulated as a tuple $\left ( \mathcal{M}, \mathcal{G}, \mathcal{E}\right ) $, where $\mathcal{M}$ is the cue maker used to define what are the common cues, $\mathcal{G}$ is a graph of ``sarcasm detection process'',  $\mathcal{E}$ is cue evaluator used to determine which cues to keep selecting. 

\textbf{1. Cue maker.}  Human sarcasm judgment often relies on the combination and analysis of one or more cues to achieve an accurate understanding. Such cues can be broadly categorized into three types: linguistic cues, contextual cues and emotional cues. Linguistic cues refer to the linguistic features inherent in the text, including \textit{keywords}, \textit{rhetorical devices}, \textit{punctuation} and \textit{language style}. Contextual cues refer to the environment and background of the text, including \textit{topic}, \textit{cultural background}, \textit{common knowledge}. Emotional cues denote the emotions implied in the text, including \textit{emotional words}, \textit{special symbols (such as emojis)} and \textit{emotional contrasts}. Hence, GoC can obtain \textit{4+3+3=10} cues.

\textbf{2. Graph construction.} In $\mathcal{G}=\left ( V,E \right )$, 10 cues are regarded as vertices, constituting the vertex set $V$, the supplement relations across cues are regarded as edges. Given the cue $c_k$, the cue evaluator $\mathcal{E}$ considers cue $c_j$ to provide the most complementary information to $c_k$, which would combine with $c_k$ to facilitate a deep understanding of sarcasm.

\textbf{3. Cue evaluator.} We associate $\mathcal{G}$ with LLM detecting sarcasm process. To advance this process, the cue evaluator $\mathcal{E}$ assesses the current progress by asking the LLM whether the cumulative cues obtained thus far are sufficient to yield an accurate judgment. The search goes to an end if a positive answer is returned; otherwise, the detection process proceeds by instructing the LLM to determine which additional cues to select and in what order. In this work, an LLM will act as the cue evaluator, similar to ToT.

We employ a voting strategy to determine the most valuable cue for selection, by deliberately comparing multiple potential cue candidates in a voting prompt, such as:

\noindent \colorbox{gray!13}{\parbox{7.5cm}{Given an input text $\mathcal{X}$, the target is to accurately detect sarcasm. Now, we have collected the keyword information as the first step: \{\textit{keywords}\}, judge if this provides over 95\% confidence for accurate detection. If so, output the result. Otherwise, from the remaining cues \{\textit{rhetorical devices}, \textit{punctuation}, ...\}, vote the most valuable one to improve accuracy and confidence for the next step.
}}

This step can be formulated as $\mathcal{E}\left (\mathcal{L}_{\theta}^{GoC}, c_{j+1} \right ) \sim Vote\left \{ \mathcal{L}_{\theta}^{GoC}\left ( c_{j+1}|\mathcal{X}, c_{1,2,...,j} \right )  \right \}_{c_{j+1}\in \{c_{j+1},...,c_k\}  } $. Until the final judgment is reached, the most valuable cue are always selected in a greedy fashion. Although GoC enables the exploration of many possible paths across the cue graph, its nature remains grounded in a step-by-step reasoning paradigm (see Algorithm~2 in App. A). 

\subsection{Bagging of Cues}
We relax the assumption that the cues are interrelated in detecting sarcasm. We introduce \textit{bagging of cues}, a ensemble learning based paradigm that allows LLMs to independently consider varied combinations of cues without assuming a fixed order or dependency among them (Fig.~\ref{fig:model} (c)). 

BoC constructs a pool of the pre-defined 10 cues $\mathcal{C}$. From this pool, $\mathcal{T}$ subsets are obtained through $\mathcal{T}$ random samplings, where each subset $\mathcal{S}_{t}$ consists of $q~\left (i.e., 1\le q \le 10 \right ) $ cues. BoC thus leverages LLMs to generate $\mathcal{T}$ independent sarcasm predictions $\hat{y}_t$  based on the cues of each subset. Finally, such predictions are aggregated using a majority voting mechanism to produce the final result. This approach embraces randomness in cue selection, enhancing the LLM's ability to explore numerous potential paths. 
BoC consists of three key steps:

\textbf{Step 1.} Cue subsets construction. A total of $\mathcal{T}$ cue subsets $\mathcal{S}_{t \in [1, 2, ..., \mathcal{T}]}=\left \{ c_{t1}, c_{t2},...,c_{tq} \right \}$ are created by randomly sampling without replacement from the complete pool of cues $\mathcal{C}$. Each sampling is independent.

\textbf{Step 2.} LLM prediction. For each subset $\mathcal{S}_t$, a LLM $\mathcal{L}_{\theta}^{BoC}$ is used to independently make sarcasm prediction through the comprehensive analysis of the cues in the subset and the input text. This can be conceptually encapsulated as $\hat{y}_t \sim \mathcal{L}_{\theta } ^{BoC}\left ( \hat{y}_t| \mathcal{S}_t, \mathcal{X} \right ) $.

\textbf{Step 3.} Prediction aggregation. Such predictions $\{ \hat{y}_1, \hat{y}_2,...,\hat{y}_{\mathcal{T} } \}$ are then combined using majority voting to yield the final prediction: $Y$.

BoC does not follow the step-by-step reasoning paradigm for sarcasm detection (see Algorithm~3 in App. A.) 

\subsection{Tensor of Cues}
CoC and GoC methods mainly handle low-order interactions between cues, while BoC assumes cues are independent. To capture high-order interactions among cues, we introduce \textit{tensor of cues}, a stereo paradigm that allows LLMs to amalgamate three types of cues ($viz.$ linguistic, contextual and emotional cues) into a high-dimensional representation. (Fig.~\ref{fig:model} (d)).

ToC treats each type of cues as an independent, orthogonal view for sarcasm understanding, and constructs a multi-view representation through the tensor product of such three types of cues. We first ask the LLM to extract linguistic, contextual, and emotional cues respectively via a simple prompt. For example:  


\noindent \colorbox{gray!13}{\parbox{7.5cm}{Extract the linguistic cues from the input sentence for sarcasm detection, such as keywords, rhetorical devices, punctuation and language style.}}

We take the outputs of the LLM's final hidden layer as the embeddings of the linguistic, contextual and emotional cues, and apply a tensor fusion mechanism to fuse the cues as additional inputs to the sarcasm detection prompt. Inspired by the success of tensor fusion network (TFN) for multi-modal sentiment analysis~\cite{zadeh-2017-tensor}, we apply token-wise tensor fusion to aggregate the cues. In particular, the embeddings are projected on a low-dimensional space via the fully-connected layers, i.e., $\vec{Lin} =\left ( e_1^{l}, e_2^{l},...,e_L^{l} \right )^T $, $\vec{Con} =\left ( e_1^{c}, e_2^{c},...,e_L^{c} \right )^T $, $\vec{Emo} =\left ( e_1^{e}, e_2^{e},...,e_L^{e} \right )^T $. Then, a tensor product is computed to combine the cues into a high-dimensional representation $\mathcal{Z} = \left ( e_1, e_2,...,e_L \right )^T$, where
\begin{equation} 
\begin{aligned}
e_i = 
\begin{bmatrix}
e_i^{l}\\
1 
\end{bmatrix}
 \otimes 
\begin{bmatrix}
e_i^{c}\\
1 
\end{bmatrix} 
\otimes 
\begin{bmatrix}
e_i^{e}\\
1 
\end{bmatrix}, \forall i \in [1,2,...,L].
\end{aligned}
\end{equation} 
The additional value of 1 facilitates an explicit rendering of single-cue features and bi-cue interactions, leading to a comprehensive fusion of different cues encapsulated in each fused token $e_i \in \mathcal{R} ^{(d_l+1) \times (d_c+1) \times (d_e+1)}$. The values of $d_l$, $d_c$ and $d_e$ are delicately chosen such that the dimensionality of fused token is precisely $d$\footnote{Otherwise the fused tokens are truncated to d-dim vectors}. That enables an integration of the aggregated cues to the main prompt via:

\noindent \colorbox{gray!13}{\parbox{7.5cm}{Consider the information provided in the current cue above. Classify whether the input text is sarcastic or not. If you think the Input text is sarcastic, answer: yes. If you think the Input text is not sarcastic, answer: no.}}

\noindent The embedded prompt above is \textbf{prepended} with the aggregated cue sequence $\mathcal{Z}$ before fed to the LLM. As it is expected to output a single token of ``yes'' or ``no'' by design, we take the logit of the first generated token and decode the label accordingly as the output of ToC. 

ToC facilitates deep interactions among these cues (see Algorithm~4 in App. A). Notably, as ToC manipulates cues on the vector level via neural structures, it requires access to the LLM structure and calls for supervised training on a collection of labeled samples. During training, the weights of the LLM are frozen, and the linear weights in $f_{lin}, f_{con}, f_{emo}$ are updated as an adaptation of LLM to the task context.

\section{Experiments}
\subsection{Experiment Setups}\label{sec:setups}
\textbf{Datasets.} Four benchmarking datasets are selected as the experimental beds, $viz.$ IAC-V1~\cite{lukin-walker-2013-really}, IAC-V2~\cite{oraby-etal-2016-creating},  SemEval 2018 Task 3~\cite{van-hee-etal-2018-semeval} and MUStARD~\cite{mustard}. 
The details and statistics for each dataset are shown in Table~1 in App. B.

\noindent \textbf{Baselines.} A wide range of SOTA baselines are included for comparison. They are: 
\begin{itemize}[itemsep=0pt, topsep=0pt, parsep=4pt, leftmargin=*]
	\item{\textbf{Prompt tuning.}~ (1) \underline{\textbf{IO}}, (2) \underline{\textbf{CoT}}~\cite{wei2022chain} and (3) \underline{\textbf{ToT}}~\cite{yao2024tree} are three SOTA prompting approaches by leveraging advanced prompt approaches to enhance LLM's performance.}
	\item{\textbf{LLMs.} We involve four general LLMs in the experiment, including (4) \underline{\textbf{GPT-4o}}, (5) \underline{\textbf{Claude 3.5 Sonnet}}, (6) \underline{\textbf{Llama 3-8B}} and (7) \underline{\textbf{Qwen 2-7B}}~\cite{qwen}. The first two are non-open-source LLMs while the last two are open-source LLMs. All four LLMs are representative of the strongest capabilities of their kinds.} 
\end{itemize}

\noindent \textbf{Implementation.} We have implemented the prompting methods for \textbf{GPT-4o}, \textbf{Claude 3.5 Sonnet}, \textbf{Llama 3-8B} and \textbf{Qwen2-7B}. The GPT-4o and Claude 3.5 Sonnet methods are implemented with the respective official Python API library: openAI\footnote{https://github.com/openai/openai-python} and anthropic\footnote{https://github.com/anthropics/anthropic-sdk-python}, while the LLaMA and Qwen methods are implemented based on the Hugging Face Transformers library\footnote{https://huggingface.co/docs/transformers}. Further details are presented in App. C.

\begin{table*}[t]
\centering
\caption{Performance on four datasets. For LLMs, all strategies are based on a zero-shot setting. \textbf{\textcolor{blue}{Blue}} and \textcolor{purple}{purple} indicate the best and second-best results for each dataset. $\clubsuit$ represents significance improvement over the best baseline via unpaired t-test (p $<$ 0.05).}
\label{tab:baseline}
\small
\scalebox{0.85}{
\begin{tabular}{clcccccccc>{\columncolor{pink!20}}c}
\toprule
\multirow{2}{*}{\textbf{Paradigm}} & \multirow{2}{*}{\textbf{Method}} & \multicolumn{2}{c}{\textbf{IAC-V1}}                    & \multicolumn{2}{c}{\textbf{IAC-V2}}                       & \multicolumn{2}{c}{\textbf{SemEval 2018}}                  & \multicolumn{2}{c}{\textbf{MUStARD}}    & \multirow{2}{*}{\textbf{Avg. of F1}}                   \\ \cline{3-10}
                 &                & Acc.                      & Ma-F1                     & Acc.                      & Ma-F1                     & Acc.                      & Ma-F1                     & Acc.                      & Ma-F1        \\ \midrule[1pt]







 \multirow{6}{*}{\textbf{GPT-4o}} 
& IO                        & 70.63     &  70.05   & \textcolor{purple}{\underline{73.03}}                       & \textcolor{purple}{\underline{71.99}}                     &  64.03                         & 63.17 &  67.24    & 65.79        &   67.75         \\

&   CoT      & 61.56                    &  58.49                     &58.83                        & 56.42                    &  58.92                         & 51.99                       &  58.11    & 55.76        &  55.67         \\

&   ToT      & \textcolor{purple}{\underline{71.56}}                    &  \textcolor{purple}{\underline{71.17}}                     & 70.63                        & 69.07                     &  63.90                         & 63.02                       &  69.00    & 68.27       &  \textcolor{purple}{\textbf{\underline{67.88}}}          \\

\cdashline{2-10}

&  \textbf{CoC (Ours)}      & \textbf{\textcolor{blue}{\underline{72.19}}}                    &  \textbf{\textcolor{blue}{\underline{71.52}}}                     & \textbf{\textcolor{blue}{\underline{73.36}}}                        & \textbf{\textcolor{blue}{\underline{72.31}}}                    &  \textcolor{purple}{\underline{70.79}}                         & \textcolor{purple}{\underline{70.60}}                       &  \textcolor{purple}{\underline{69.42}}    & \textcolor{purple}{\underline{68.48}}        &  \textcolor{blue}{\textbf{\underline{70.73}}}$^{\clubsuit}$         \\

&  \textbf{GoC (Ours)}       & 65.00                    &  62.91                     & 64.97                        & 61.30                     &  \textbf{\textcolor{blue}{\underline{74.03}}}$^{\clubsuit}$                         & \textbf{\textcolor{blue}{\underline{74.02}}}$^{\clubsuit}$                       &  \textbf{\textcolor{blue}{\underline{70.69}}}$^{\clubsuit}$    & \textbf{\textcolor{blue}{\underline{69.91}}}$^{\clubsuit}$        &  67.04        \\

&  \textbf{BoC (Ours)}       & 68.75                    &  67.36                     & 71.35                       & 69.39                     &  62.12                         & 61.85                       &  \textcolor{purple}{\underline{69.42}}    & 68.45 &  66.76          \\

\midrule[1pt]

\multirow{6}{*}{\textbf{Claude 3.5 Sonnet}} 
& IO                        & 66.56                    &  66.54                     & \textbf{\textcolor{blue}{\underline{76.78}}}                        & \textbf{\textcolor{blue}{\underline{76.62}}}                     &  75.13                         & 75.11                       &  \textbf{\textcolor{blue}{\underline{74.78}}}    & \textbf{\textcolor{blue}{\underline{74.78}}}        &  \textcolor{purple}{\textbf{\underline{73.26}}}          \\

&   CoT      & \textbf{\textcolor{blue}{\underline{71.25}}}                    &  \textbf{\textcolor{blue}{\underline{71.14}}}                     & 74.66                        & 74.10                     &  71.56                         & 71.47                       &  73.62    & 73.53        &  72.56          \\

&   ToT      & 63.44                    &  62.48                     & 71.88                        & 71.74                     &  68.62                         & 68.61                       &  58.84    & 54.46       &  64.32          \\

\cdashline{2-10}

& \textbf{CoC (Ours)}      & 69.69                    &  69.40                     &  73.22                       & 73.17                     &  \textbf{\textcolor{blue}{\underline{82.27}}}$^{\clubsuit}$                         &  \textbf{\textcolor{blue}{\underline{82.23}}}$^{\clubsuit}$                      & \textcolor{purple}{\underline{74.20}}     & \textcolor{purple}{\underline{74.16}}        &  \textcolor{blue}{\textbf{\underline{74.74}}}$^{\clubsuit}$          \\

&   \textbf{GoC (Ours)}       & \textcolor{purple}{\underline{70.94}}                    &  \textcolor{purple}{\underline{70.93}}                     & \textcolor{purple}{\underline{74.67}}                       & \textcolor{purple}{\underline{74.18}}                     &  \textcolor{purple}{\underline{76.91}}                         & \textcolor{purple}{\underline{76.91}}                       &  70.00    & 69.85        &  72.97         \\

&  \textbf{BoC (Ours)}       & 66.88                    &  66.40                     & 73.61                        & 72.82                     &  70.28                         & 70.07                      &  72.61    & 71.93 &  70.31           \\

\midrule[1pt]

\multirow{7}{*}{\textbf{Llama 3-8B}} 
& IO                        & 55.94 & 46.40 & 54.70 & 43.74 & 49.36 & 44.46 & 54.64 & 44.99 & 44.90       \\
 
&   CoT      & 56.25 & 47.28 & 54.22 & 42.96 & 49.36 & 44.55 & 54.20 & 44.86 & 44.91         \\

&   ToT     & 52.50 & 48.98 & 55.95 & 53.05 & 50.64 & 48.63 & 54.35 & 50.56 &  50.31 \\

\cdashline{2-10}

&  \textbf{CoC (Ours)}      & 56.25 & 46.95 & 54.03 & 42.60 & 49.23 & 44.36 & 54.93 & 45.66 & 44.89     \\

&   \textbf{GoC  (Ours)}       & 57.10                    &  54.96                     & 54.22                        & 53.30                    &  57.33    & 57.24 &  52.77    & 52.67        & 54.54          \\

&  \textbf{BoC (Ours)}       & \textbf{\textcolor{blue}{\underline{62.50}}}                    &  \textcolor{purple}{\underline{59.28}}                     & \textcolor{purple}{\underline{62.57}}                        & \textcolor{purple}{\underline{58.11}}                     &  \textcolor{purple}{\underline{65.94}}                         & \textcolor{purple}{\underline{65.50}}                     & \textcolor{purple}{\underline{59.71}}                           &  \textcolor{purple}{\underline{56.70}}                     &   \textcolor{purple}{\textbf{\underline{59.90}}}        \\

&   \textbf{ToC (Ours)}       & \textcolor{purple}{\underline{62.19}}       & \textbf{\textcolor{blue}{\underline{61.78}}}$^{\clubsuit}$        &  \textbf{\textcolor{blue}{\underline{72.95}}}$^{\clubsuit}$      &  \textbf{\textcolor{blue}{\underline{72.94}}}$^{\clubsuit}$      & \textbf{\textcolor{blue}{\underline{68.88}}}$^{\clubsuit}$           & \textbf{\textcolor{blue}{\underline{68.21}}}$^{\clubsuit}$       & \textbf{\textcolor{blue}{\underline{61.26}}}$^{\clubsuit}$         & \textbf{\textcolor{blue}{\underline{58.03}}}$^{\clubsuit}$  & \textcolor{blue}{\textbf{\underline{65.24}}}$^{\clubsuit}$  \\

\midrule[1pt]

\multirow{7}{*}{\textbf{Qwen 2-7B}} 
& IO                        & \textcolor{purple}{\underline{56.56}} & \textcolor{purple}{\underline{49.32}} & 51.82 & 38.57 & 45.15 & 38.83 & 54.78 & 46.17 & 43.22
         \\

&   CoT      & 54.69 & 46.53 & 52.88 & 40.12 & 43.24 & 35.79 & \textcolor{purple}{\underline{54.93}} & 45.81 & 42.06
      \\

&   ToT      & 53.44                    &  43.71                     & 50.29                        & 39.62                     &  44.26                         & 38.12                       &  52.90    & 44.60       &  41.51          \\

\cdashline{2-10}

&  \textbf{CoC (Ours)}      & 55.00 & 45.77 & 51.92 & 38.90 & 43.75 & 36.37 & 53.77 & 44.26 & 41.33         \\

&   \textbf{GoC (Ours)}       & 55.00                    &  47.35                     & \textcolor{purple}{\underline{53.45}}                        & \textcolor{purple}{\underline{42.25}}                     &  45.03                         & 38.17                       &  54.49    & \textcolor{purple}{\underline{47.49}}        &  43.82         \\

&  \textbf{BoC (Ours)}       & 52.50                    &  43.78                     & 52.40                        & 40.24                     &  \textcolor{purple}{\underline{49.87}}                         & \textcolor{purple}{\underline{45.63}}                       &  54.06    & 46.11 &  \textcolor{purple}{\textbf{\underline{43.94}}}          \\

&   \textbf{ToC (Ours)}       & \textbf{\textcolor{blue}{\underline{71.56}}}$^{\clubsuit}$                    &  \textbf{\textcolor{blue}{\underline{71.56}}}$^{\clubsuit}$                     & \textbf{\textcolor{blue}{\underline{72.33}}}                        & \textbf{\textcolor{blue}{\underline{71.76}}}$^{\clubsuit}$                     &  \textbf{\textcolor{blue}{\underline{68.88}}}$^{\clubsuit}$                         & \textbf{\textcolor{blue}{\underline{68.77}}}$^{\clubsuit}$                       &  \textbf{\textcolor{blue}{\underline{65.94}}}$^{\clubsuit}$    & \textbf{\textcolor{blue}{\underline{61.46}}}$^{\clubsuit}$ &  \textcolor{blue}{\textbf{\underline{68.39}}}$^{\clubsuit}$          \\

\bottomrule[1pt]
\end{tabular}
}
\end{table*}

\subsection{Main Results} \label{Main Results}

We report both \textbf{Accuracy} and \textbf{Macro-F1} scores for \textbf{SarcasmCue} and baselines in Table~\ref{tab:baseline}. 


\textbf{(1) SarcasmCue consistently outperforms SoTA prompting baselines.} The proposed prompting strategies in the \textbf{SarcasmCue} framework achieve an overall superior performance compared to the baselines and consistently push the SoTA by 4.2\%, 2.0\%, 29.7\% and 58.2\% on F1 scores across four datasets. 
In particular, by explicitly designing the reasoning steps for sarcasm detection, CoC beats CoT by a tremendous margin on \textbf{GPT-4o} and \textbf{Claude 3.5 Sonnet}, whilst performing in par with CoT on \textbf{Llama 3-8B} and \textbf{Qwen 2-7B}. By pre-defining the set of cues in three main categories, GoC and BoC effectively guide LLMs to reason along correct paths, leading to more accurate judgments of sarcasm compared to the freestyle thinking in ToT. For example, the best proposed method, CoC (74.74), brings a 2.0\% improvement over the best baseline method, IO (73.26).
ToC achieves an effective tensor fusion of multi-aspect cues for sarcasm detection, significantly outperforming other baselines. For instance, it exhibits a 29.7\% improvement over the best baseline method, ToT (50.31).

\textbf{(2) Sarcasm detection does not necessarily follow a step-by-step reasoning process.} The comparison between sequential (CoT, CoC, GoC, ToT) and non-sequential (BoC, ToC) prompting strategies fails to provide clear empirical evidences on whether sarcasm detection follows a step-by-step reasoning process. Nevertheless, the results on \textbf{Llama 3-8B} are more indicative to \textbf{GPT-4o} and \textbf{Claude 3.5 Sonnet}, since the latter models have strong capabilities on their own (IO) and do not significantly benefit from any prompting strategies. For \textbf{Llama 3-8B} and \textbf{Qwen 2-7B}, non-sequential methods, particularly ToC, show superior performance. In \textbf{Llama 3-8B}, ToC achieves an average F1 score of 65.24\%, which is 8.9\% higher than the best sequential method (GoC at 54.54\%). The difference is even more pronounced on \textbf{Qwen 2-7B}. This seems to support our hypothesize that sarcasm has a non-sequential nature.

\begin{table*}[t]
\centering
\caption{Ablation study of BoC, GoC and ToC. All strategies are run on a zero-shot setting. The best results for each dataset are colored in \textcolor{blue}{blue}. }  
\label{tab:ablation}
\footnotesize
\scalebox{0.85}{
\begin{tabular}{clcccc>{\columncolor{pink!20}}c}
\toprule
\textbf{LLMs}  & \textbf{Method}  & \textbf{IAC-V1} & \textbf{IAC-V2} & \textbf{SemEval}                                                             & \textbf{MUStARD}  & \textbf{Avg. of F1}\\
\toprule
\multirow{8}{*}{\textbf{Claude 3.5}}
& \textit{w/o} Lin    & 68.41     & \textbf{\textcolor{blue}{\underline{75.62}}}              &77.42           & 69.66         & 72.78       \\
&\textit{w/o} Emo    & 69.65     & 74.04              &\textbf{\textcolor{blue}{\underline{78.70}}}           & \textbf{\textcolor{blue}{\underline{70.57}}}          & \textcolor{blue}{\textbf{\underline{73.24}}}        \\

& \textit{w/o} Con  & 70.53     & 74.91              &76.39           & 70.11         & 72.99     \\
& \textbf{GoC}  &  \textbf{\textcolor{blue}{\underline{70.93}}}                & 74.18               & 76.91      & 69.85        &  72.97    \\

\cline{2-7}
& \textit{w/o} Lin    & 45.89     & 42.49              &47.47           & 65.33      & 50.30          \\
& \textit{w/o} Emo    & 58.00     & 56.99              &56.81           & 68.84       & 60.16           \\

& \textit{w/o} Con  & 61.71     & 63.70              &69.53           & 74.80    &   67.44       \\
&  \textbf{BoC}  & \textbf{\textcolor{blue}{\underline{66.40}}}                     & \textbf{\textcolor{blue}{\underline{72.82}}}                     & \textbf{\textcolor{blue}{\underline{70.07}}}      & \textbf{\textcolor{blue}{\underline{71.93}}} &  \textcolor{blue}{\textbf{\underline{70.31}}}           \\
\cline{2-7}

\toprule 
\toprule

\multirow{12}{*}{\textbf{Llama 3-8B}}
& \textit{w/o} Lin    & 45.79  & 51.90   & 56.01   & 46.84   & 50.14 \\
&\textit{w/o} Emo    & 48.60     & 49.40              &52.38           & 45.12   & 48.88                \\
& \textit{w/o} Con  & 52.51     & \textbf{\textcolor{blue}{\underline{53.69}}}              &52.14           & 48.28 & 51.66             \\
& \textbf{GoC}  & \textbf{\textcolor{blue}{\underline{54.96}}}     & 53.30              &\textbf{\textcolor{blue}{\underline{57.24}}}           & \textbf{\textcolor{blue}{\underline{52.67}}}          & \textbf{\textcolor{blue}{\underline{54.54}}}    \\

\cline{2-7}

& \textit{w/o} Lin    & 52.71     & 57.51              &57.53           & 53.06          & 55.20     \\
& \textit{w/o} Emo    & 57.33     & 59.40              &62.01           & 53.06              & 57.95    \\

& \textit{w/o} Con  & 56.88     & \textbf{\textcolor{blue}{\underline{60.36}}}              &59.04           & 52.30        & 57.15      \\
&  \textbf{BoC}  &  \textbf{\textcolor{blue}{\underline{59.28}}}                        & 58.11                         & \textbf{\textcolor{blue}{\underline{65.50}}}                          &  \textbf{\textcolor{blue}{\underline{56.70}}}                     &   \textcolor{blue}{\textbf{\underline{59.90}}}             \\

\cline{2-7}
&   \textit{w/o} Lin    & 53.31    & 67.05             & 59.20          & 48.05   & 56.90  \\
&  \textit{w/o} Emo    & 57.42    & 67.08             & 64.01          &  52.89        &  60.35   \\
&  \textit{w/o} Con  &  55.26    &  71.78            &  63.93         &  52.48         & 60.86 \\
&  \textbf{ToC}  & \textbf{\textcolor{blue}{\underline{61.78}}}    &  \textbf{\textcolor{blue}{\underline{72.94}}}            & \textbf{\textcolor{blue}{\underline{68.21}}}          & \textbf{\textcolor{blue}{\underline{58.03}}}          & \textcolor{blue}{\textbf{\underline{65.24}}}    \\

\toprule 

\end{tabular}
}
\end{table*}
\subsection{Ablation Study}
Table~\ref{tab:ablation} presents the result of ablation study. \textit{w/o Lin}, \textit{w/o Emo}, \textit{w/o Con} refer to the method where linguistic, emotional and contextual cues are ablated, respectively. To avoid proactive extraction of ablated cues by an LLM, we explicitly ``prompt away'' the cues in the inputs. An example prompt could be ``You can only use the emotional cues and contextual cues, and do not use any linguistic information here'' for the \textit{w/o Lin} case.

The experiment results highlight the following conclusions: (a) the removal of any single type of cue leads to a noticeable drop in performance across all datasets, demonstrating the importance of each type of cue in sarcasm detection; (b) linguistic cues appear to have the most significant impact, as removing them leads to a noticeable decrease in performance across most settings; (c) the absence of contextual cues also affects the performance, but to a lesser extent compared to linguistic cues.

\subsection{Zero-shot v/s Few-shot Prompting}\label{sec:fewshot}
\begin{figure}[t]
    \centering
    \includegraphics[width=2.86in]{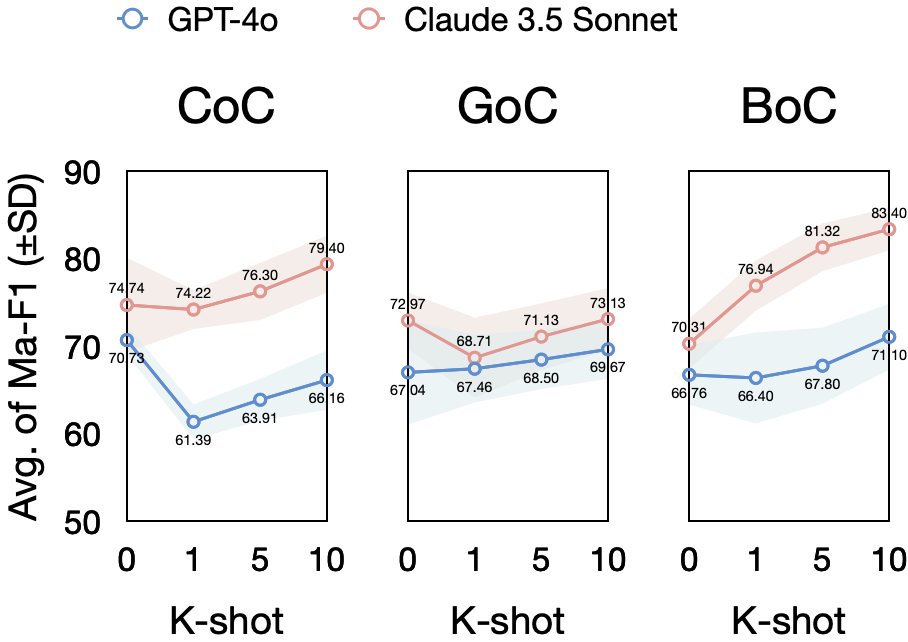}
  \caption{The average Macro-F1 across K-shots for the \textbf{GPT-4o} and \textbf{Claude 3.5 Sonnet} models.}
   \label{fig:k-shotss}
\end{figure}

Since the above experiments are mainly based on a zero-shot setting, we are curious of whether the conclusions also apply in a few-shot scenario. Therefore, we perform few-shot experiments to evaluate whether the proposed SarcasmCue framework can perform better when a limited number of contextual examples are available. We plot the main results in Fig.~\ref{fig:k-shotss}, we randomly sample $k= \left \{ 0,1,5,10 \right \} $ examples from the training set. Please refer to Table~2, App. D for the full result. 

As shown in the plot, the number of demonstrations has a significant impact on the results. For example, CoC appears sensitive to the initial introduction of demonstration examples with a slight descent in performance when only 1 example is provided. However, as the number of shots increases to 5 and 10, the performance progressively improves. This trend underscores the effectiveness of CoC in adapting and refining its approach with more examples. In contrast, BoC demonstrates a consistent improvement in performance as the number of shots increases. 

Overall, these results demonstrate the robustness and adaptability of the SarcasmCue framework in zero-shot and few-shot scenarios. The framework can effectively utilize limited contextual examples to further improve sarcasm detection, making it suitable for applications where large annotated datasets are not readily available.

\begin{figure}[t]
    \centering
    \includegraphics[width=2.85in]{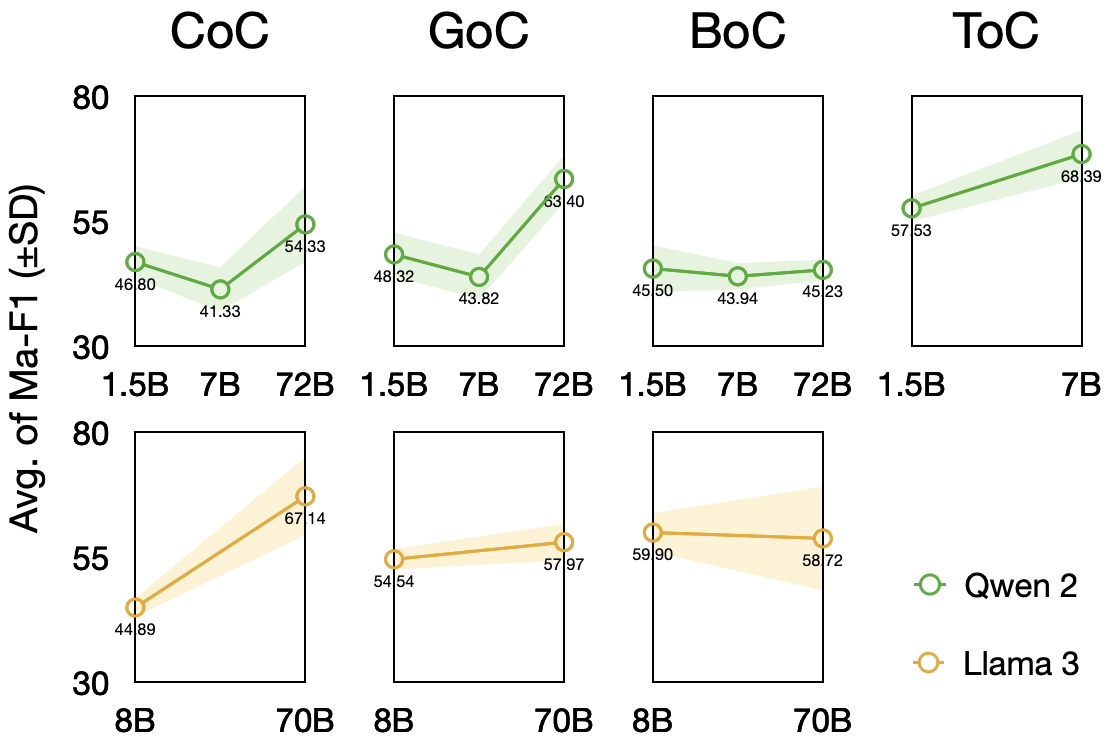}
  \caption{The influence of model scale. The figures in the top and bottom correspond to Qwen and Llama models, respectively.}
   \label{fig:model_scale}
\end{figure}

\subsection{Influences of LLM scales}

In an attempt to study the influence of different LLM scales, we evaluate the performance of sarcasm detection of \textbf{Qwen} and \textbf{Llama} of varying sizes, see Fig.~\ref{fig:model_scale}. 

The key take-aways are two-fold. First, the efficacy of our prompting methods is amplified with increasing model scale. This aligns closely with the key findings of the CoT method~\cite{wei2022chain}. This occurs because when an LLM is sufficiently large, its capabilities for multi-hop reasoning and understanding language are significantly enhanced. Second, ToC exhibits high sensitivity to model scale, performing significantly better in larger models, making it particularly suitable for larger-scale applications.
CoC and GoC demonstrate moderate sensitivity, indicating a balance between performance improvement and scalability.
BoC offers robust performance even in smaller models, suggesting its utility in resource-constrained scenarios.
Overall, our proposed framework has a high adaptability across various model scales by offering suitable methods. Please see Table 3 and Fig. 1, App. E for the full results.

\subsection{Error Analysis}\label{sec:error}

\begin{figure}[t]
    \centering
    \includegraphics[width=2.3in]{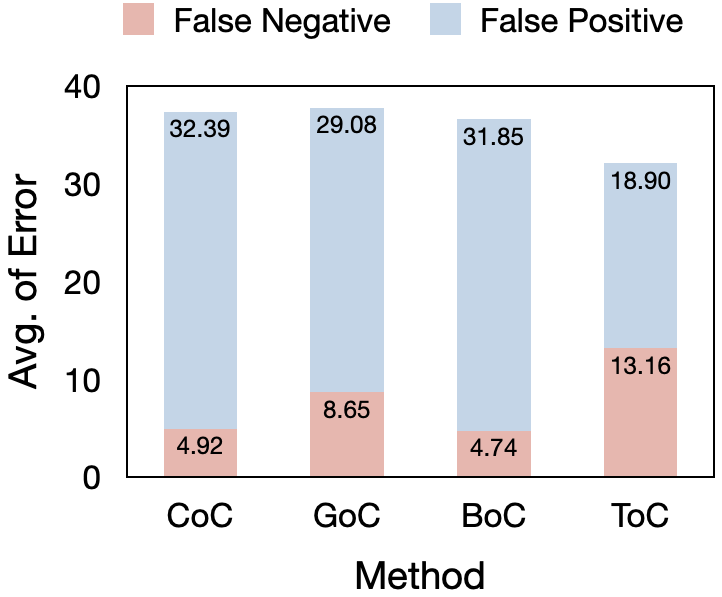}
  \caption{The average error rate of the four prompting methods.}
   \label{fig:error_rate}
\end{figure}

Fig.~\ref{fig:error_rate} shows the error rates of failure cases in terms of false negative (FN) and false positive (FP) for all four prompting methods in SarcasmCue. 
CoC, GoC and BoC exhibit higher false positive rates, indicating an over-detection of sarcasm that could lead to the frequent misclassification of normal statements as sarcastic. In contrast, ToC exhibits the lowest overall error rate and the FP and FN rates are indeed much closer to each other, indicating a balanced performance in detecting both sarcastic and non-sarcastic texts. These insights highlight potential directions for future improvements in sarcasm detection methodologies. The higher false positive rates suggest a need for refining these methods to reduce over-sensitivity and improve discrimination between sarcastic and non-sarcastic texts. The detailed case study is presented in App. F.


\subsection{Extension to New Task}
To evaluation the generalization capability of SarcasmCue, we apply it to another complex affection understanding task, \textbf{humor detection}. We compare our proposed SarcasmCue (where the backbone is GPT-4o) with two supervised PLMs (MFN~\cite{Hasan_Lee_Rahman_Zadeh_Mihalcea_Morency_Hoque_2021} and SVM+BERT~\cite{zhang2024cmma}) on two benchmarking datasets, CMMA~\cite{zhang2024cmma} and UR-FUNNY-V2~\cite{hasan2019ur}. 

As shown in Table~\ref{tab:humor}, our methods (BoC and CoC) surpass the baseline on CMMA, whilst performing in par to the strongest baselines on the UR-FUNNY-V2 dataset. These results highlight the strong generalizability and versatility of our framework, confirming its potential utility across a wide range of affection understanding tasks.

\begin{table}[t]
\centering
\footnotesize
\caption{Performance on two humor detection datasets.} 
\label{tab:humor}
\scalebox{0.85}{
\begin{tabular}{ccccc>{\columncolor{pink!20}}c}
\toprule
 \multirow{2}{*}{\textbf{Method}} & \multicolumn{2}{c}{\textbf{CMMA}}   &  \multicolumn{2}{c}{\textbf{UR-FUNNY-V2}} &\multirow{2}{*}{\textbf{Avg. of F1}}   \\ \cline{2-5}
& Acc.                      & Ma-F1                     & Acc.                      & Ma-F1 \\
\toprule
\textbf{MFN} &  -  &  -  &  64.44  &  64.12  &  -  \\
 \textbf{SVM+BERT} &  55.23  &  54.08  &  \textbf{\textcolor{blue}{\underline{69.62}}}  &  \textbf{\textcolor{blue}{\underline{69.27}}}  &  61.68  \\
\toprule
\textbf{CoC} &  78.14  &  \textbf{\textcolor{blue}{\underline{58.60}}}  &  64.08  &  60.13  &  65.24  \\
\textbf{GoC} &  79.60  &  57.42  &  64.89  &  61.65  &  65.89  \\
\textbf{BoC} &  \textbf{\textcolor{blue}{\underline{75.81}}}  &  58.58  &  68.71  &  66.83  &  \textbf{\textcolor{blue}{\underline{67.48}}}  \\
       \toprule  
\end{tabular}
}
\end{table}
\section{Conclusions}
This work aims to study the stepwise reasoning nature of sarcasm detection, and introduces a prompting framework (called SarcasmCue) containing four sub-methods, $viz.$ CoC, GoC, BoC and ToC. It elicits LLMs to detect human sarcasm by considering sequential and non-sequential prompting methods. 
Our comprehensive evaluations across multiple benchmarks and SoTA LLMs demonstrate that SarcasmCue outperforms traditional methods and pushes the state-of-the-art by 4.2\%, 2.0\%, 29.7\% and 58.2\% F1 scores across four datasets. Additionally, the performance of SarcasmCue on humor detection further validate its robustness and versatility.

\textbf{Limitations.} 
SarcasmCue has its limitation: it incorporates only three types of cues, while other potentially useful cues have not been integrated, potentially limiting the model's comprehensive understanding of sarcasm.

\bibliography{tacl2021}
\bibliographystyle{acl_natbib}

\iftaclpubformat

\onecolumn

\appendix
\section{A. Algorithms of Four Prompting Methods}
\label{app:algorithm}
\textbf{1. CoC.} We present further details of CoC in Algorithm~\ref{alg:coc}.

\begin{algorithm}[htp]
    \caption{Chain of contradiction}
    \label{alg:coc}
    \small
    \begin{algorithmic}[1]
        \Require
            \State \textbf{Input:} Sentence \(\mathcal{X}\), an LLM $\mathcal{L}_{\theta}$
        
        \Ensure
            \State \textbf{Output:} Sarcasm Label \(\mathcal{Y}\)
        
        \State \textbf{Step 1:} Detect surface sentiment
            \State Output cue \( c_1 \): \( c_1 \sim \mathcal{L}_{\theta}^{CoC}\left ( c_1| \mathcal{X}, p_1 \right ) \)
        
        \State \textbf{Step 2:} Discover true intention
            \State Output cue \( c_2 \): \( c_2 \sim \mathcal{L}_{\theta}^{CoC}\left ( c_2| \mathcal{X}, c_1, p_2 \right ) \)
        
        \State \textbf{Step 3:} Evaluate consistency and make prediction
            \State Output cue \( c_3 \): \( c_3 \sim \mathcal{L}_{\theta}^{CoC}\left ( c_3| \mathcal{X}, c_1, c_2, p_3 \right ) \)
            \State \( \mathcal{Y} = \begin{cases} \text{Sarcastic} & \text{if } c_1 \neq c_2 \\ \text{Not Sarcastic} & \text{otherwise} \end{cases} \)
        
        \State \Return \( \mathcal{Y} \)
    \end{algorithmic}
\end{algorithm}

\textbf{2. GoC.} We present further details of GoC in Algorithm~\ref{alg:goc}.

\begin{algorithm}[htp]
    \caption{Graph of Cues (GoC) for Sarcasm Detection}
    \label{alg:goc}
    \small
    \begin{algorithmic}[1]
        \Require
            \State \textbf{Input:} Sentence \(\mathcal{X}\), an LLM $\mathcal{L}_{\theta}$
        
        \Ensure
            \State \textbf{Output:} Sarcasm Label \(\mathcal{Y}\)

        
        \State \textbf{1. Graph Construction}
            \State Construct graph \(\mathcal{G} = (V, E)\) where 10 cues are vertices \(V\) and relationships between cues are edges \(E\)
        
        \State \textbf{2. Sarcasm Detection Process}
            \State Initialize selected cues \(C_{\text{selected}} = \emptyset\), \(j=0\)
            \State Initialize current confidence \( \mathbb{C}  = 0\)

            \While{$ \mathbb{C} < 0.95 \cap  j\le 10$}
                \State Select the most valuable cue:
                \State \( c_{\text{j+1}} \sim Vote\left \{ \mathcal{L}_{\theta}^{GoC}\left ( c_{j+1}|\mathcal{X}, c_{1},c_2,...,c_j \right )  \right \}_{c_{j+1}\in \{c_{j+1},...,c_{10}\}  } \)
                \State Add \( c_{\text{j+1}} \) to \( C_{\text{selected}} \)
                \State Update current confidence \( \mathbb{C} \), $j$++
            \EndWhile
        
            \State Make final judgment based on \( C_{\text{selected}} \):
                \( \mathcal{Y} = \mathcal{L}_{\theta}^{GoC}\left ( \mathcal{Y}|\mathcal{X}, C_{\text{selected}} \right ) \)

        \State \Return \( \mathcal{Y} \)
        
    \end{algorithmic}
\end{algorithm}

\textbf{3. BoC.} We present further details of BoC in Algorithm~\ref{alg:boc}.

\begin{algorithm}[hpt]
    \caption{Bagging of cues}
    \label{alg:boc}
    \small
    \begin{algorithmic}[1]
        \Require
            \State \textbf{Input:} Sentence \(\mathcal{X}\), Cue Pool \(\mathcal{C}\), Number of Subsets \(\mathcal{T}\), Number of Cues per Subset \(q\),  an LLM $\mathcal{L}_{\theta}$
        
        \Ensure
            \State \textbf{Output:} Sarcasm Label \(Y\)
        
        \State \textbf{Step 1: Cue Subsets Construction}
            \For{$t = 1$ \textbf{to} $\mathcal{T}$}
                \State Randomly sample a subset \(\mathcal{S}_t = \{c_{t1}, c_{t2}, \ldots, c_{tq}\}\) from \(\mathcal{C}\)
            \EndFor
        
        \State \textbf{Step 2: LLM Prediction}
            \For{$t = 1$ \textbf{to} $\mathcal{T}$}
                \State Generate sarcasm prediction \(\hat{y}_t \sim \mathcal{L}_{\theta}^{BoC}(\hat{y}_t | \mathcal{S}_t, \mathcal{X})\)
            \EndFor
        
        \State \textbf{Step 3: Prediction Aggregation}
            \State Aggregate predictions using majority voting:
            \State \(Y \sim Vote(\{\hat{y}_1, \hat{y}_2, \ldots, \hat{y}_{\mathcal{T}}\})\)
        
        \State \Return \(Y\)
        
    \end{algorithmic}
\end{algorithm}

\textbf{4. ToC.} We present further details of ToC in Algorithm~\ref{alg:toc}.

\begin{algorithm}[!t]
    \caption{Tensor of cues}
    \label{alg:toc}
    \small
    \begin{algorithmic}[1]
        \Require
            \State \textbf{Input:} Sentence \(\mathcal{X}\), an LLM $\mathcal{L}_{\theta}$
        
        \Ensure
            \State \textbf{Output:} Sarcasm Label \(\mathcal{Y}\)

        \State \textbf{Step 1: Extract Cues}
            \State Obtain linguistic cue embeddings \( \vec{Lin} = (e_1^l, e_2^l, \ldots, e_m^l)^T \), contextual cue embeddings \( \vec{Con} = (e_1^c, e_2^c, \ldots, e_p^c)^T \), emotional cue embeddings \( \vec{Emo} = (e_1^e, e_2^e, \ldots, e_s^e)^T \)

        \State \textbf{Step 2: Construct Tensor Representation}
            \State Compute tensor product to combine cues: \(\mathcal{Z} = \begin{bmatrix} \vec{Lin}\\ 1 \end{bmatrix} \otimes \begin{bmatrix} \vec{Con}\\ 1 \end{bmatrix} \otimes \begin{bmatrix} \vec{Emo}\\ 1 \end{bmatrix} \)

        \State \textbf{Step 3: Sarcasm Detection}
            \State Take tensor \( \mathcal{Z} \) as input to a LLM for sarcasm detection:
            \State \( \mathcal{Y} \sim \mathcal{L}_{\theta}^{ToC}(\mathcal{Y} | \mathcal{Z}, \mathcal{X}) \)
        
        \State \Return \( \mathcal{Y} \)
        
    \end{algorithmic}
\end{algorithm}

\section{B. Datasets Details}
\textbf{Datasets.} Four benchmarking datasets are selected as the experimental beds, $viz.$ IAC-V1~\cite{lukin-walker-2013-really}, IAC-V2~\cite{oraby-etal-2016-creating},  SemEval 2018 Task 3~\cite{van-hee-etal-2018-semeval} and MUStARD~\cite{mustard}. 

\begin{table}[h]
\centering
\footnotesize
\caption{Dataset statistics.} 
\label{tab:statistics}
\scalebox{0.85}{
\begin{tabular}{ccccc}
\toprule
\textbf{Dataset}  & \textbf{Avg. Length} & \textbf{\#Train}                                                 & \textbf{\#Dev}               & \textbf{\#Test} \\
\toprule

IAC-V1        & 68              & 1,595              & 80  & 320                \\
IAC-V2       &43 & 5,216              & 262              & 1,042                  \\
SemEval 2018          & 14              &3,634            & 200  & 784               \\
MUStARD  & 14             & 552            & - & 138          \\
       \toprule  
\end{tabular}
}
\end{table}

\textbf{IAC-V1} and \textbf{IAC-V2} are from the Internet Argument Corpus (IAC)~\cite{lukin-walker-2013-really}, specifically designed for the task of identifying and analyzing sarcastic remarks within online debates and discussions. It encompasses a balanced mixture of sarcastic and non-sarcastic comments.

\textbf{SemEval 2018 Task 3} is collected using irony-related hashtags (i.e. \#irony, \#sarcasm, \#not) and are subsequently manually annotated to minimise the amount of noise in the corpuses. It emphasize the challenges inherent in identifying sarcasm within the constraints of MUStARD's concise format, and highlight the importance of context and linguistic subtleties in recognizing sarcasm.

\textbf{MUStARD} is compiled from popular TV shows including Friends, The Golden Girls, The Big Bang Theory, etc. It consists of 690 samples total of 3,000 utterances. Each sample is a conversation consisting of several utterances. In this work, we only use the textual information.

The statistics for each dataset are shown in Table~\ref{tab:statistics}.

\section{C. Implementation Details}
We have implemented the prompting methods for \textbf{GPT-4o}, \textbf{Claude 3.5 Sonnet}, \textbf{LLaMA3-8B-Instruct} and \textbf{Qwen 2-7B}. The GPT-4o and Claude 3.5 Sonnet methods are implemented with the respective official Python API library: openAI\footnote{https://github.com/openai/openai-python} and anthropic\footnote{https://github.com/anthropics/anthropic-sdk-python}, while the LLaMA and Qwen methods are implemented based on the Hugging Face Transformers library\footnote{https://huggingface.co/docs/transformers}. All prompting strategies are implemented for \textbf{GPT-4o} and \textbf{Claude 3.5 Sonnet} except for ToC, which can solely be deployed on open-sourced LLMs. 
Following previous works in this field, LangChain\footnote{https://github.com/langchain-ai/langchain} is employed for the implementation of ToT and GoC. For the training of ToC, cross-entropy loss between the output logit and the true label token is computed to update the weights of the fully-connected layers. The mean performance of each model over 5 runs is calculated. 

Given the proprietary nature of GPT-4o and Claude 3.5 Sonnet, we have implemented only CoC, GoC and BoC prompting approaches. For Llama 3-8B and Qwen 2-7B, we implemented all four proposed prompting approaches. This is due to the reasons previously discussed: ToC requires access to and modification of the base model. We run all the models on four A100 GPUs.

\begin{table*}[t]
\centering
\caption{Few shot performance testing.}\label{sec:Few-shot}
\small
\scalebox{0.85}{
\begin{tabular}{cllcccc>{\columncolor{pink!20}}c}
\toprule
\textbf{LLMs}  & \textbf{Method} & \textbf{K-shot} & \textbf{IAC-V1} & \textbf{IAC-V2} & \textbf{SemEval}                                                             & \textbf{MUStARD}  & \textbf{Avg. of F1}\\
\midrule[1pt] 
\multirow{12}{*}{\textbf{GPT-4o}} &
\multirow{4}{*}{\textbf{CoC}}&0-shot                  &\textcolor{blue}{\underline{71.52}}   &\textcolor{blue}{\underline{72.31}}    &\textcolor{blue}{\underline{70.60}}     & \textcolor{blue}{\underline{68.48}}  &\textcolor{blue}{\underline{70.73}}  \\ 
&&1-shot                  &58.91   &61.99   &63.69     & 60.98  &61.39  \\ 
&&5-shot                  &60.65   &66.00   &65.39     & 63.60  &63.91  \\ 
&&10-shot                 &63.34   &70.29   &67.59     & 63.40  &66.16  \\ \cline{2-8}

&\multirow{4}{*}{\textbf{GoC}}&0-shot                 &  62.91  &  61.30  &  \textcolor{blue}{\underline{74.02}}  & \textcolor{blue}{\underline{69.91}}        &  67.04\\ 
&&1-shot                  & 66.00  &  \textcolor{blue}{\underline{66.70}}  &  73.00  &  64.12  &  67.45\\ 
&&5-shot                  & 66.93  &  65.88  &  73.41  &  67.77  &  68.50\\ 
&&10-shot                 & \textcolor{blue}{\underline{74.38}}  &  66.36  &  69.34  &  68.61  &  \textcolor{blue}{\underline{69.67}}\\ \cline{2-8}
            
& \multirow{4}{*}{\textbf{BoC}}&0-shot                  &67.36   &\textcolor{blue}{\underline{69.39}}   &61.85     & 68.45  &66.76  \\ 
&&1-shot                  &64.67   &66.36   &61.08     & 73.47  &66.40  \\ 
&&5-shot                  &65.66   &67.52   &64.06     & 73.97  &67.80  \\ 
&&10-shot                 &\textcolor{blue}{\underline{70.70}}   &69.06   &\textcolor{blue}{\underline{68.12}}   & \textcolor{blue}{\underline{76.51}}   &\textcolor{blue}{\underline{71.10}}  \\ 

\midrule[1pt] 

\multirow{12}{*}{\textbf{Claude 3.5 Sonnet}} &
\multirow{4}{*}{\textbf{CoC}}&0-shot 
&69.40   &73.17   &82.23     & 74.16     &74.74     \\  
&&1-shot                   &72.45   &74.38   &77.29     & 72.75  & 74.22\\ 
&&5-shot                  &71.99   &78.12   &79.42     & 75.65  & 76.30\\ 
&&10-shot                 &\textcolor{blue}{\underline{75.49}}   &\textcolor{blue}{\underline{79.07}}   &\textcolor{blue}{\underline{83.46}}     & \textcolor{blue}{\underline{79.56}}  & \textcolor{blue}{\underline{79.40}}\\  \cline{2-8}

&\multirow{4}{*}{\textbf{GoC}}&0-shot                 &  70.93  &  74.18  &  \textcolor{blue}{\underline{76.91}}  &  \textcolor{blue}{\underline{69.85}}        &  72.97\\ 
&&1-shot                  &  65.16  & 69.29  &  74.98   & 65.40  & 68.71\\ 
&&5-shot                  &  69.01  & 72.76  &  75.62   & 67.12  & 71.13\\ 
&&10-shot                 &  \textcolor{blue}{\underline{72.80}}  & \textcolor{blue}{\underline{74.71}}  &  76.65   & 68.36  & \textcolor{blue}{\underline{73.13}}\\ \cline{2-8}

&\multirow{4}{*}{\textbf{BoC}}&0-shot 
&66.40   &72.82   &70.07     & 71.93     & 70.31     \\ 
&&1-shot                   &74.63   &81.09   &76.64     & 75.38  & 76.94\\ 
&&5-shot                  &78.40   &84.34   &79.72     & 82.83  & 81.32\\ 
&&10-shot                 &\textcolor{blue}{\underline{80.27}}   &\textcolor{blue}{\underline{84.82}}   &\textcolor{blue}{\underline{82.76}}     & \textcolor{blue}{\underline{85.76}}  & \textcolor{blue}{\underline{83.40}}\\  \midrule[1pt] 
\end{tabular}}
\end{table*}

\section{D. Zero-shot v/s Few-shot Prompting}\label{sec:fewshot2}
We perform zero-shot and few-shot experiments to evaluate whether the proposed SarcasmCue framework can perform better when a limited number of contextual examples are available. The results are shown in Table~\ref{sec:Few-shot}. We design four $k$-shot settings: zero-shot, one-shot, five-shot, ten-shot. For each setting, we randomly sample $k= \left \{ 0,1,5,10 \right \} $ examples from the training set. 

The impact of adding shots varies with the number of shots. For example, CoC appears sensitive to the initial introduction of demonstration examples with a slight descent in performance when only 1 example is provided. However, as the number of shots increases to 5 and 10, the performance progressively improves. This trend underscores the effectiveness of CoC in adapting and refining its approach with more examples. In contrast, BoC demonstrates a consistent improvement in performance as the number of shots increases. Compared to CoC and BoC, GoC exhibits a relatively lower sensibility to the presence of demonstration examples, while still showing a slight but stable improvement with more shots.

Overall, these results demonstrate the robustness and adaptability of the SarcasmCue framework in zero-shot and few-shot scenarios. The framework can effectively utilize limited contextual examples to improve sarcasm detection, making it suitable for applications where large annotated datasets are not readily available. This adaptability underscores the practical value of SarcasmCue in real-world settings where training data may be scarce.



\begin{table*}[t!]
\centering
\caption{Influence of model scale. Macro-F1 score is measured on all four datasets, and the average Macro-F1 score is computed and shown in the last column.} 
\label{tab:model_scale}
\small
\scalebox{0.85}{
\begin{tabular}{clcccc>{\columncolor{pink!20}}c}
\toprule
\textbf{LLMs}  & \textbf{Method}  & \textbf{IAC-V1} & \textbf{IAC-V2} & \textbf{SemEval}                                                             & \textbf{MUStARD}  & \textbf{Avg. of F1}\\
\toprule

\multirow{4}{*}{\textbf{Qwen 2-1.5B}}
& \textbf{CoC}    &  48.05 & 44.43  & 44.05 & 50.66 & 46.80\\
&\textbf{GoC}     & 43.75 & 53.21 & 50.69 & 45.63 & 48.32\\
& \textbf{BoC}    & 43.84 & 42.86 & 42.87 & 52.41 & 45.49\\
& \textbf{ToC}    & 57.46 & 57.60 & 60.69 & 54.40 & 57.53\\

\toprule 

\multirow{4}{*}{\textbf{Qwen 2-7B}}
& \textbf{CoC}    & 45.77 & 38.90 & 36.37 & 44.26 & 41.33\\
&\textbf{GoC}     & 47.35 & 42.25 & 38.17 & 47.49 & 43.82\\
& \textbf{BoC}    & 43.78 & 40.24 & 45.63 & 46.11 & 43.94\\
& \textbf{ToC}    & 71.56 & 71.76 & 68.77 & 61.46 & 68.39\\
\toprule 

\multirow{3}{*}{\textbf{Qwen 2-72B}}
& \textbf{CoC}  & 61.33  & 59.42  & 44.92 & 51.63 & 54.33    \\
&\textbf{GoC}     & 57.67 &68.78 & 65.28 & 61.87 & 63.40\\
& \textbf{BoC}     & 45.24  & 43.38  & 44.18 & 48.10 & 45.23 \\
\toprule 
\multirow{3}{*}{\textbf{LlaMA 3-8B}}
& \textbf{CoC}    & 46.95 & 42.60 & 44.36  & 45.66 & 44.89\\
&\textbf{GoC}     & 54.96 & 53.30 & 57.24 & 52.67 & 54.54 \\
& \textbf{BoC}    & 59.28 & 58.11 & 65.50 &  56.70 & 59.90 \\

\toprule 

\multirow{3}{*}{\textbf{LlaMA 3-70B}}
& \textbf{CoC}    & 68.94 & 77.29 & 62.59 & 59.73 & 67.14  \\
&\textbf{GoC}     & 56.83 & 62.57 & 58.66 & 53.81 &  57.97\\
& \textbf{BoC}    & 65.49 & 68.52 & 55.14 & 45.72 & 58.72 \\

\toprule 

\end{tabular}
}
\end{table*}
\begin{figure*}[t!]
    \centering
    \includegraphics[width=6in]{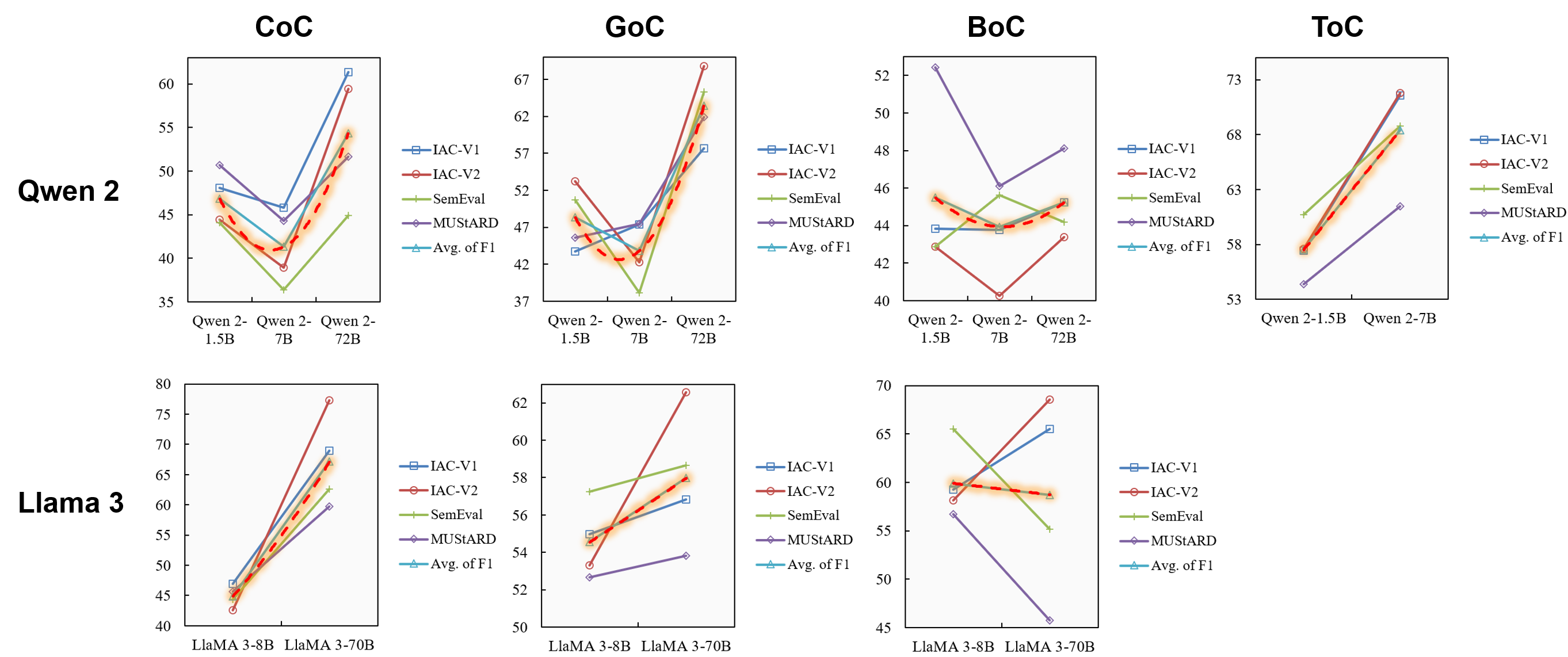}
  \caption{The influence of model scale.}
   \label{fig:model_scale_2}
\end{figure*}

\section{E. Influences of LLM scales}
In an attempt to study the influence of different LLM scales, we evaluate the performance of sarcasm detection of Qwen and Llama of varying sizes. Table~\ref{tab:model_scale} presents the macro-F1 scores of each model across the four sarcasm detection tasks.

The key take-aways are two-fold. First, with increasing model scale, the efficacy of our prompting is exponentially amplified. This aligns closely with the key findings of the CoT method~\cite{wei2022chain}. This is because when an LLM is sufficiently large, its capabilities for multi-hop reasoning are greatly developed and strengthened. More specifically:

(1) CoC demonstrates a significant improvement in performance as model scale increases. For Qwen models, the average F1 score rises from 46.80\% (1.5B) to 54.33\% (72B). LLaMA models show an even more pronounced enhancement, with the average F1 score jumping from 44.89\% (8B) to 67.14\% (70B). This indicates that CoC becomes more effective with larger model scales.

(2) GoC also exhibits a positive trend with increasing model size. In Qwen models, performance improves from 48.32\% (1.5B) to 63.40\% (72B) average F1 score. LLaMA models display a similar trend, with the average F1 score increasing from 54.54\% (8B) to 57.97\% (70B). These results suggest that GoC generally benefits from larger model scales across different architectures.

(3) BoC shows inconsistent performance across model scales. For Qwen models, performance remains relatively stable, with a slight decrease in the 72B model (45.23\% average F1) compared to smaller versions. LLaMA models demonstrate a minor decline in performance, with the average F1 score decreasing from 59.90\% (8B) to 58.72\% (70B). This suggests that BoC might be more effective with smaller model scales.

(4) ToC exhibits the most substantial improvement within the available data range. For Qwen models, the average F1 score increases dramatically from 57.53\% (1.5B) to 68.39\% (7B). 

Overall, our proposed framework demonstrates high adaptability across different model scales by offering a range of methods. This adaptability allows for optimized performance based on available computational resources and specific task requirements

\begin{table*}[t]
\centering
\small
\caption{Typical examples for case study.} 
\label{tab:case_study}
\scalebox{0.85}{
\begin{tabular}{clccccc}
\toprule
\textbf{Example}  &  \textbf{Text}  &  \textbf{Golden}  &  \textbf{CoC}  &  \textbf{GoC}  &  \textbf{BoC}  &  \textbf{ToC}  \\

\toprule
1 & 
\noindent \colorbox{gray!13}{\parbox{8.2cm}{Now that is funny, the marie troll not knowing its a troll.}}
& Sarcastic & \faCheckSquare & \faCheckSquare &  \faCheckSquare  & \faCheckSquare  \\

2 & 
\noindent \colorbox{gray!13}{\parbox{8.2cm}{You are aware that words have more than one meaning, right? And that every definition isn't appropriate in every situation? The definition, from dictionary.com, that you should have used is:
To infer or estimate by extending or projecting known information.
}}
& Sarcastic & \faTimes & \faTimes &  \faCheckSquare  & \faCheckSquare  \\

3 & 
\noindent \colorbox{gray!13}{\parbox{8.2cm}{Do you grasp the concept of ``consentual''?
consentual definition | Dictionary.com
}}
& Sarcastic & \faTimes & \faTimes &  \faTimes  & \faCheckSquare  \\

4 & 
\noindent \colorbox{gray!13}{\parbox{8.2cm}{No, this is the point of the 10th amendment. Article 1 Section 8 applies to Congress...the 10th amendment grants all powers not listed to the states or people. 
The 14th amendment is not the ``federal government can do whatever'' amendment.
}}
& Sarcastic & \faTimes & \faTimes &  \faTimes  & \faTimes  \\

5 & 
\noindent \colorbox{gray!13}{\parbox{8.2cm}{You make it seem as if you are doing me a favor by reading what I post
}}
& Sarcastic & \faCheckSquare & \faTimes &  \faCheckSquare  & \faCheckSquare  \\

6 & 
\noindent \colorbox{gray!13}{\parbox{8.2cm}{Just out of interest, which particular aspect of ``truth'' are you getting at here?
}}
& Sarcastic & \faCheckSquare & \faCheckSquare &  \faTimes  & \faCheckSquare  \\

7 & 
\noindent \colorbox{gray!13}{\parbox{8.2cm}{You forgot to mention that we would have to change our numbering system so that grasshoppers had 4 legs.
}}
& Sarcastic & \faCheckSquare & \faCheckSquare & \faCheckSquare & \faTimes  \\

\toprule 

8 & 
\noindent \colorbox{gray!13}{\parbox{8.2cm}{Science is the current sum of human knowledge about how the world works.
}}
& Not Sarcastic & \faCheckSquare & \faCheckSquare &  \faCheckSquare  & \faCheckSquare  \\

9 & 
\noindent \colorbox{gray!13}{\parbox{8.2cm}{I think its actually the states job...the judiciary does need to overturn Roe v. Wade to get this done though...which doesn't mean it becomes illegal.
}}
& Not Sarcastic & \faTimes & \faCheckSquare &  \faCheckSquare  & \faCheckSquare  \\

10 & 
\noindent \colorbox{gray!13}{\parbox{8.2cm}{Mmmmm, not necessarily. Many of the arguments of against gods (those with specific properties, not just a general diety) deal with incompatible traits, like a square circle has. One does not have to search the universe to know square circles do not exist.People state simple negatives all the time. The lack of evidence for the positive makes them reasonable.
}}
& Not Sarcastic & \faTimes & \faTimes &  \faCheckSquare  & \faCheckSquare  \\

11 & 
\noindent \colorbox{gray!13}{\parbox{8.2cm}{Apples and oranges. We're not demanding that they have abortions either.
}}
& Not Sarcastic & \faTimes & \faTimes &  \faTimes  & \faCheckSquare  \\

12 & 
\noindent \colorbox{gray!13}{\parbox{8.2cm}{and how do you know this......oh I see...you said ``I think''....but you don't really ``know'' what most Americans favor or don't favor...you just "think"
}}
& Not Sarcastic & \faTimes & \faTimes &  \faTimes  & \faTimes  \\

13 & 
\noindent \colorbox{gray!13}{\parbox{8.2cm}{
Well, there certainly is here with these cats, because they're not actually inheriting a trait; the symptoms are being independently induced in all the cats, parents and offspring, by denying them all particular nutrients.
}}
& Not Sarcastic & \faCheckSquare & \faTimes & \faCheckSquare & \faCheckSquare  \\

14 & 
\noindent \colorbox{gray!13}{\parbox{8.2cm}{The human collective is the authority. One major advantage of this authority over a theistic one is that it actually exists.
}}
& Not Sarcastic & \faCheckSquare & \faCheckSquare & \faTimes & \faCheckSquare  \\

15 & 
\noindent \colorbox{gray!13}{\parbox{8.2cm}{Did you read the article? A capuchin is type of monkey, in this case, the type that was used in the experiment.
}}
& Not Sarcastic & \faCheckSquare & \faCheckSquare & \faCheckSquare & \faTimes  \\

\toprule
\end{tabular}
}
\end{table*}

\section{F. Case Study} 
We analyze the proposed four prompting approaches on several typical cases in Table~\ref{tab:case_study}. We categorize and analyze sarcasm detection methods. In scenarios involving straightforward statements (Examples 8, 15), all methods correctly identify texts as non-sarcastic, showcasing the SarcasmCue framework's efficacy in clear-cut non-sarcastic contexts. For scenarios marked by clear linguistic contrasts (Examples 1, 6, 7), the CoC and GoC methods demonstrate superior performance. They effectively capture textual contradictions, making them ideally suited for texts where the apparent meaning sharply diverges from the intended message.

For texts involving complex contexts that necessitate an understanding of nuanced background knowledge (Examples 2, 3, 9, 10), the BoC and ToC methods prove more effective. BoC achieves this through sampling multiple subsets of cues, thus capturing the complexity of the context, whereas ToC employs a multi-view representation to process intricate high-order interactions.

In scenarios characterized by subtle sarcasm (Examples 5, 11)—where texts may lack overt sarcastic markers or structural clues—ToC outperforms other methods. It excels in capturing the intricate interaction among linguistic, contextual, and emotional cues. Additionally, for texts involving specialized domain knowledge (Examples 13, 14), both BoC and ToC are effective due to their ability to integrate and analyze domain-specific cues.

This analysis highlights that different sarcasm detection methods are tailored to specific textual scenarios. CoC and GoC are highly effective in environments with straightforward linguistic oppositions, where the sarcasm is direct and easily discernible. Conversely, BoC and ToC are particularly adept in scenarios that demand a deeper understanding of complex and subtle cues. ToC is especially notable for its performance across a broad range of scenarios, attributed to its capability to capture and analyze complex interactions among multiple layers of cues.

However, in highly ambiguous situations, a blend of methods or the addition of extra contextual information may be required. This insight directs future research towards identifying or combining the most appropriate methods for enhancing the overall accuracy of sarcasm detection across varied scenarios.

\end{document}